\pdfoutput=1
\documentclass[conference]{IEEEtran}
\IEEEoverridecommandlockouts

\usepackage[T1]{fontenc}
\usepackage{cite}
\usepackage{amsmath}
\usepackage{mathptmx}   
\usepackage{graphicx}
\usepackage{booktabs}
\usepackage{array}
\usepackage{textcomp}
\usepackage{url}
\usepackage{microtype}

\begin{document}

\title{Belief-Space Perception Routing under Coupled Sensor Faults and Compute Contention}

\author{
\IEEEauthorblockN{Sparsh Roy\textsuperscript{1,2}, Vihan Aggarwal\textsuperscript{3}, Davin Yin\textsuperscript{4}}
\IEEEauthorblockA{%
\textsuperscript{1}\textit{Massachusetts Institute of Technology}, Cambridge, MA, USA\\
\textsuperscript{2}\textit{Hopewell Valley Central High School}, New Jersey, USA\\
\textsuperscript{3}\textit{Millburn High School}, New Jersey, USA\\
\textsuperscript{4}\textit{Harvard University}, Cambridge, MA, USA\\
sparshr@mit.edu, vihanaggarwal@gmail.com, davin\_yin@college.harvard.edu}
}

\maketitle

\begin{abstract}
A robot that has to see and react on a fixed clock runs into two problems at once. Its cameras degrade in rain, mud, fog, and darkness. And the single onboard processor it runs on is shared with planning and control, so the compute left over for perception moves around from second to second. Most systems model the two separately. We present a perception router that tracks probabilistic estimates of sensor-fault state and compute-contention state, couples them with a noisy-OR term, and uses the coupled estimate to pick one of four detector configurations (YOLO11x/n at 1280 or 640 px) so that the frame finishes before its deadline. Where the two stressors co-occur, the coupled policy cuts the deadline-miss rate by 1.1 to 9.4 percentage points against a policy that treats them independently. The interval excludes zero in five of six conditions, the pooled effect over 10 sequences and 6 conditions has sign-test $p$ = 0.001, and every uncoupled control and the fault-free trajectory sit at exactly 0.0 pp. Routing costs tens of microseconds per frame. We then asked whether the coupling the method exploits arises on its own. Across eight real RADIATE adverse-weather sequences and three workload proxies independent of the fault signal, after Benjamini-Hochberg correction and a replication run, none of 24 tests found it. We report that null and scope the routing result as a proof of mechanism. Whether such coupling occurs in the field is still open, and the released evaluation pipeline lets a deployment settle it on its own traces.
\end{abstract}

\begin{IEEEkeywords}
autonomous systems, robot perception, sensor fault detection, compute contention, deadline-constrained inference, adaptive inference, hidden Markov model
\end{IEEEkeywords}

\section{Introduction}

Mud on a lens, rain streaks, fog, a night exposure that washes out detail: a robot working outdoors loses image quality in ways that are hard to predict and hard to ignore. The compute is a second problem: one small onboard processor usually serves perception, planning, and control at the same time, so the share available to a detector moves around from second to second. A perception stack that ignores either effect will sometimes hand the planner an answer that is late, or wrong, or both.

The two problems are usually modeled separately, and they need not be independent. Rough terrain that throws mud onto a lens is also terrain that makes a controller work hard, and a sudden loss of visibility can set off extra localization work. How often that happens in a deployed system is an empirical question, and Section~\ref{sec:coupling} takes it up. Where it does happen, a router that treats the two signals as unrelated can commit to an expensive detector configuration at the moment the processor has least room for it, or go on trusting a degraded camera because evidence of the fault has not yet reached the decision.

Existing work handles each stressor on its own. Fault-tolerant perception monitors down-weight or flag degraded sensor streams [1], [2]. Compute-aware inference picks a cheaper network path when a latency budget tightens [3], [4]. We are not aware of a system that estimates both as latent states and uses the relationship between them.

We contribute (i) a coupled belief state built from a per-channel switching hidden Markov model (HMM) for sensor fault and a two-state HMM for compute contention, both fit without task-specific labels; (ii) a noisy-OR routing policy with one coupling coefficient $\kappa$, calibrated on held-out data, that adds tens of microseconds per frame; (iii) an evaluation on three real robotics datasets covering two camera geometries, with five seeds and 95\% confidence intervals, where the negative results are reported alongside the positive ones; and (iv) a direct test of the coupling assumption the method rests on, using workload proxies independent of the fault signal under a multiple-testing correction. Code, data splits, and results are at github.com/VihanAggarwal/belief-space-perception-routing.

\begin{figure}[t]
\centering
\includegraphics[width=\columnwidth]{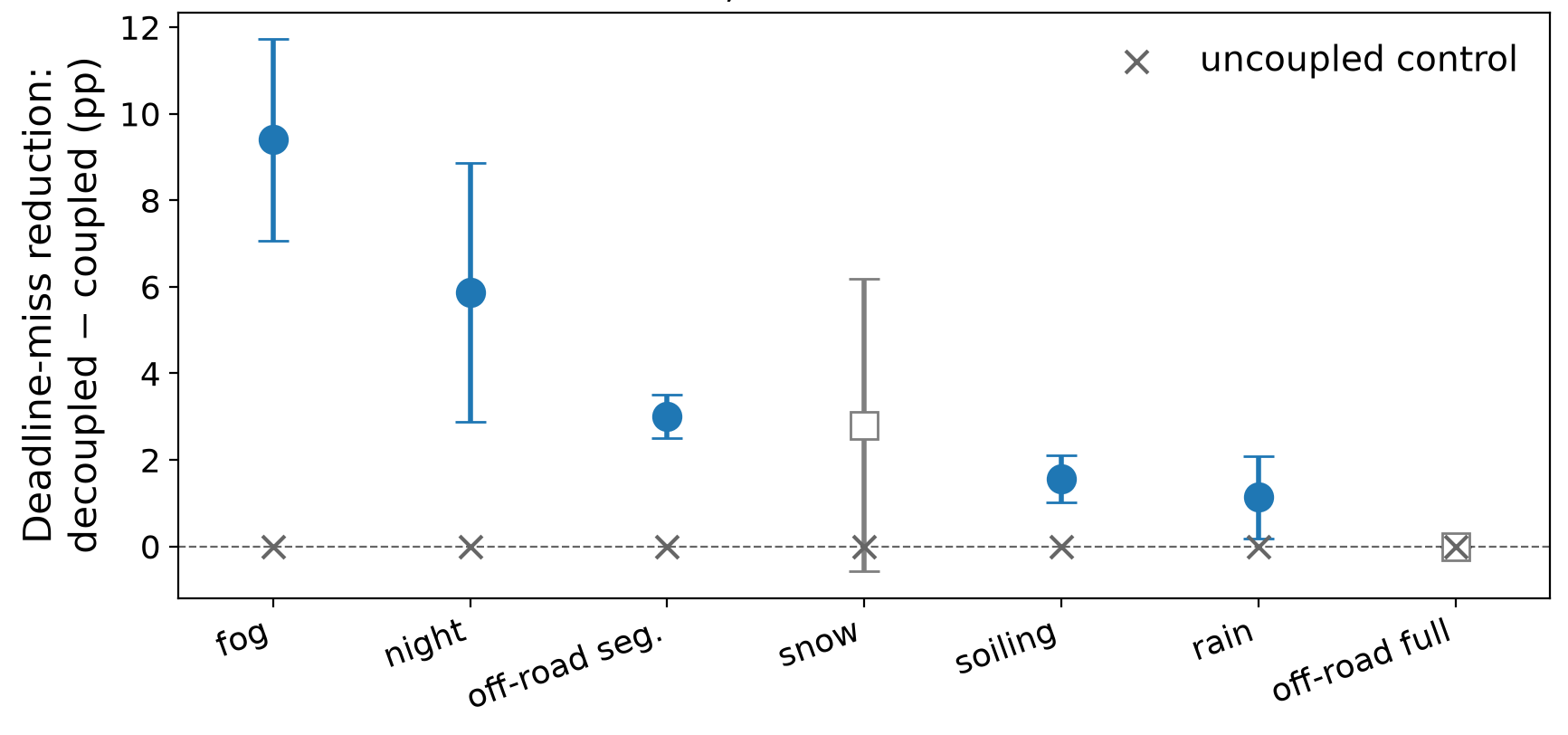}
\caption{Deadline-miss reduction of the coupled policy over the decoupled baseline under imposed, fault-conditioned contention, with 95\% intervals over 5 seeds. Filled circles: coupling enabled, interval excludes zero. Open square: coupling enabled, not significant (snow). Crosses: matched uncoupled control, exactly 0.0 pp. TartanDrive appears twice, as a fault-free full trajectory and as an off-road segment with real faults.}
\label{fig:rqh}
\end{figure}

Fig.~\ref{fig:rqh} previews the main result. The coupled policy misses fewer deadlines in every condition that contains real faults, by 1.1 pp on rain and by 9.4 pp on fog, and the reduction is exactly zero in every uncoupled control and on the fault-free trajectory.

\section{Related Work}

TartanDrive [5] provides large-scale off-road driving data, RADIATE [6] real adverse-weather sequences, and WoodScape [7] fisheye frames with real lens-soiling masks. Between them they cover two camera geometries and several kinds of degradation. Sensor-fault monitoring is usually deterministic or binary [1]. Stress-testing studies [2] and the RoboDrive Challenge [8] show detection accuracy collapsing under per-stream corruption, but none of this work models a continuous fault belief coupled to anything else. Compute-aware inference treats the compute budget as a constant or a per-frame observable, whether the mechanism is an early-exit network [3], [4] or a driving-specific exit policy [9]; the budget is never a latent state estimated jointly with sensor health. Adaptive fusion [10], dynamic multimodal fusion [11], and LLM-guided modality routing [12] gate modalities without modeling contention at all. We know of no prior work that couples a temporally smoothed sensor-fault belief to a compute-contention belief for deadline-constrained routing, or that tests whether the coupling is present in the data it is evaluated on.

\section{Method}

\textbf{Configuration frontier.} The router chooses among four detector configurations ordered by cost: C1, YOLO11x at 1280 px; C2, YOLO11x at 640; C3, YOLO11n at 1280; C4, YOLO11n at 640. C1 is the reference. Task accuracy for the others is IoU-matched detection agreement (F1) with C1 on the same frame, a pseudo-ground-truth whose validity Section~\ref{sec:coupling} examines. The ordering is consistent across datasets, and C1 through C4 give up 0.24 to 0.53 F1 in exchange for latency. We set the deadline per device to the median end-to-end C1 latency under nominal load, 211 to 341 ms on our hardware. C1 is then borderline feasible when uncontended and infeasible under contention, while the cheaper configurations still fit. Calibrating the deadline this way creates routing pressure without hand-tuning and keeps the comparison device-agnostic.

\begin{figure}[t]
\centering
\includegraphics[width=\columnwidth]{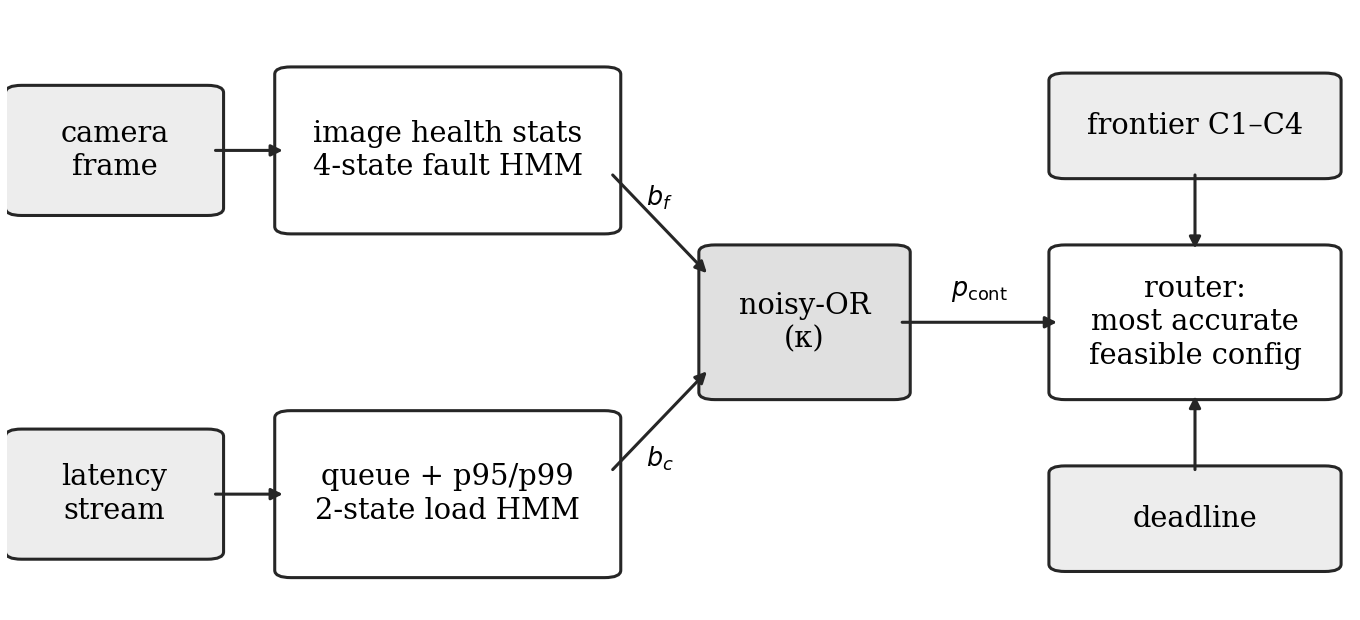}
\caption{System overview. Cheap per-frame image statistics drive a four-state sensor-fault HMM, and recent latencies drive a two-state contention HMM. The noisy-OR of (\ref{eq:noisyor}) combines the two beliefs into $p_{\mathrm{cont}}$, and the router picks the most accurate configuration that still meets the deadline under the predicted compute state. The decoupled baseline sets $p_{\mathrm{cont}} = b_c$ and drops the $b_f$ term.}
\label{fig:overview}
\end{figure}

\textbf{Sensor-fault belief.} Each image channel gets a four-state switching HMM (nominal, degrading, faulted, recovering) with Gaussian emissions and transitions fit from labeled segments. The observations are cheap image statistics: blur from Laplacian variance, polar-unwrapped for fisheye; illumination from histogram entropy; and occlusion from the survival rate of sparse optical-flow tracks, or from the real soiling-mask fraction on the fisheye data. Forward filtering gives a per-channel $b_{f,k}^{t} = P(z_k^t = \mathrm{faulted} \mid o_{1:t})$, and the frame-level belief is $b_f^{t} = \max_k b_{f,k}^{t}$. On real RADIATE rain the occlusion channel reaches balanced accuracy 0.95. The trackers recover injected faults at balanced accuracy 0.85 to 0.89 under controlled injection with known onsets, with a 3 to 6 frame lag, and every channel correctly reports no sustained fault on a fault-free trajectory. These detection numbers are scored against independent ground truth, never against the pseudo-GT above.

\textbf{Compute-contention belief.} A two-state HMM (nominal, contended) runs over the p95, p99, and queue depth of recent frame latencies. We induce contention with device-appropriate background load, which shifts p95 latency by roughly 2.5$\times$. Forward filtering gives $b_c^{t} = P(z_c^t = \mathrm{contended} \mid \ell_{1:t})$.

\textbf{Coupled policy.} Both policies pick, among the configurations feasible under the predicted compute state (median latency in that state meets the deadline), the one with the highest expected accuracy. The only difference is how the contention probability is formed. The decoupled baseline sets $p_{\mathrm{cont}}^{t} = b_c^{t}$. The coupled policy combines the two beliefs with a noisy-OR:

\begin{equation}
p_{\mathrm{cont}}^{t} = 1 - (1 - b_c^{t})(1 - \kappa\, b_f^{t})
\label{eq:noisyor}
\end{equation}

We fit $\kappa$ on a held-out calibration draw from the same trace, never the test window; it lands near 0.80 across datasets. At $\kappa$ = 0 the coupled policy reduces exactly to the decoupled one. The decoupled baseline still uses $b_f$ for accuracy weighting, so the whole difference between the two policies comes from the coupling term. We also compare against a threshold rule, $p_{\mathrm{cont}} = \max(b_c, \mathbf{1}[b_f > 0.5])$; a memoryless detector with no HMM smoothing; an oracle router handed the true contention state; and a reactive-latency baseline that responds only to observed latency and never sees the camera.

\section{Experimental Setup}

Two questions organize the results. RQ-H: when the two stressors are coupled, does coupling-aware routing miss fewer deadlines than a decoupled baseline? RQ-C: does that coupling occur in un-manipulated data, and does the benefit carry across sequences? A smaller third question, RQ-A, concerns what temporal smoothing of the fault signal buys.

Table~\ref{tab:rqh} lists the evaluated conditions with their frame counts and fault fractions alongside the headline result. TartanDrive appears twice, once as a fault-free full trajectory that serves as a negative control and once as a 309-frame rough-terrain segment with real degradation. WoodScape supplies real soiling-mask ground truth for occlusion in place of the optical-flow proxy. RADIATE covers four weather conditions whose fault fractions run between 1.9\% and 44.4\%, which lets us ask whether the benefit tracks how often faults occur or how badly they bite. A separate injection track, built by applying point-spread-function blur, exposure crush, and composited occluders to real frames with known onsets, only measures belief-tracker accuracy and dwell behavior; no headline number depends on it.

We impose the coupling between sensor fault and contention experimentally. In the coupled regime the contention schedule is conditioned on the fault state, which yields $r$ = 0.31 to 0.76 depending on the condition. The matched uncoupled control delivers the same amount of contention on a periodic schedule, and $r$ there sits between $-$0.12 and 0.01; that control is the falsification test for every coupled condition.

All runs are on an Apple M5 (Metal/MPS, fp32). The deadline self-calibrates per device, so the deadline-miss comparison does not depend on absolute latencies, which are development-environment numbers with no standing of their own. Every comparison uses five seeds and a 95\% $t$-interval on the paired difference. The real frames are identical across seeds, so seed variance reflects sensor-measurement noise on one fixed trajectory; unless stated otherwise, ``significant'' means that interval excludes zero. Two other resampling units appear, and we name them where they are used: a moving-block bootstrap within a single trace, and a cluster bootstrap over independent sequences, which is the only unit that speaks to generalization across trajectories. Repeated tests across sequences and proxies use Benjamini-Hochberg correction.

\begin{table*}[t]
\caption{Evaluated conditions and headline result. Deadline-miss rate, coupled vs. decoupled routing, under imposed coupling.}
\label{tab:rqh}
\centering
\footnotesize
\setlength{\tabcolsep}{6pt}
\begin{tabular}{lrrcccc}
\toprule
\textbf{Condition} & \textbf{Frames} & \textbf{Fault \%} & \textbf{$\kappa$} & \textbf{Coupled miss} & \textbf{Decoup. miss} & \textbf{$\Delta$ (pp), 95\% CI} \\
\midrule
Off-road, full (clean) & 1198 & 0.0 & n/a & n/a & n/a & 0.00 \\
Off-road, segment & 309 & 30.4 & 0.79 & 0.249 & 0.279 & +3.00 [2.49, 3.51]* \\
RADIATE rain & 2651 & 44.4 & 0.80 & 0.233 & 0.244 & +1.13 [0.18, 2.09]* \\
RADIATE snow & 2589 & 18.5 & 0.82 & 0.313 & 0.341 & +2.80 [$-$0.58, 6.18] \\
RADIATE fog & 2659 & 1.9 & 0.80 & 0.329 & 0.423 & +9.40 [7.07, 11.73]* \\
RADIATE night & 2644 & 12.7 & 0.80 & 0.313 & 0.371 & +5.87 [2.87, 8.86]* \\
WoodScape soiling & 3303 & 8.1 & 0.80 & 0.256 & 0.271 & +1.56 [1.01, 2.11]* \\
\bottomrule
\end{tabular}\\[3pt]
\begin{minipage}{\textwidth}
\scriptsize $\Delta$ = decoupled $-$ coupled miss rate in percentage points, with a 95\% $t$-interval over 5 seeds; an asterisk marks intervals that exclude zero. Off-road is TartanDrive. In the matched uncoupled control the same contention arrives on a periodic schedule, and $\Delta$ there is exactly 0.00 pp in every row. The table is computed from timing alone and uses no accuracy label.
\end{minipage}
\end{table*}

\section{Results}
\label{sec:results}

\subsection{Deadline misses under coupled stress}

Table~\ref{tab:rqh} and Fig.~\ref{fig:rqh} give the main result. The coupled policy lowers the deadline-miss rate in every condition that contains real faults, and the interval excludes zero in five of six; snow is positive but under-powered. The two uses of TartanDrive bracket the mechanism. On the fault-free full trajectory the difference is exactly 0.00 pp, while its degraded segment gives +3.00 pp at near-matched accuracy. Every matched uncoupled control is exactly 0.00 pp as well. Within the schedules we evaluated, the benefit shows up only when the coupling is there to exploit.

The size of the benefit does not follow how often faults occur or how strongly they correlate with contention. Fog has the lowest fault fraction, 1.9\%, and the weakest coupling, $r$ = 0.31, yet the largest reduction at +9.40 pp. The reason is severity: fog's episodes are rare but deep enough that the reference configuration becomes infeasible exactly when contention arrives. On fog the coupled policy gives up task accuracy to buy timeliness, 0.861 against 0.954 on pseudo-GT; accuracy is near-matched on rain, night, and soiling. The trade is tunable, since $\kappa$ controls how aggressively the router converts accuracy into timing headroom. A combined on-time utility (accuracy if the frame met its deadline, zero otherwise) is nonnegative throughout, and significantly higher for the off-road segment, rain, and night.

\subsection{Where the gain actually comes from}

Replacing the noisy-OR with a plain threshold rule, $p_{\mathrm{cont}} = \max(b_c, \mathbf{1}[b_f > 0.5])$, changes nothing we can measure. The two agree to within $\pm$0.5 pp on every condition, with all intervals covering zero, and the equivalence holds as observation noise is raised to $\sigma$ = 0.75. A small multilayer perceptron trained on the same belief features is competitive but inconsistent, better on three tracks and worse on four. The coupled feature itself carries the gain: sensor evidence informs the contention estimate, and any reasonable function of the two beliefs will do. That narrows what the paper claims. It also lowers the cost of adopting the idea, since a deployment can get the same effect from a two-line rule with no calibration.

An oracle router given the true contention state is statistically indistinguishable from the coupled policy, and in several conditions numerically worse. Ground truth does not help here because feasibility is decided at a median-latency threshold, and perfect knowledge of the current state does not by itself tighten that test. A reactive-latency baseline that never sees the camera reaches a lower miss rate in several conditions, but pays for it in accuracy, and its on-time utility is lower than the coupled policy's in five of seven conditions. Miss rate read on its own rewards a policy that sacrifices the accuracy the coupling exists to protect.

Routing overhead is small enough to ignore: 19 to 32 $\mu$s per frame as a batch-fit upper bound, at most 0.015\% of the reference configuration's latency, and 0.1 to 1.2 $\mu$s per frame in steady state. The gains survive a moving-block bootstrap ($B$ = 2000), so no short run of frames is carrying them. The fog benefit is largest where the reference configuration is borderline feasible and fades to non-significance at looser deadlines. Sweeping $\kappa$ gives a monotonic response, exactly 0.0 pp at $\kappa$ = 0 on all seven tracks and +10.9 pp at $\kappa$ = 1. A denser nine-configuration frontier leaves the fog result largely intact, +6.0 against +9.4 pp.

\subsection{Temporal smoothing}

Smoothing the fault signal with an HMM instead of thresholding it frame by frame raises detection balanced accuracy by 3 to 8 points on every faulted track, measured against independent onsets, and significantly reduces configuration switching on fog, night, and soiling. Switching goes up on rain, where fast droplet flicker gets smoothed into apparent sustained degradation, which argues for matching the smoothing time constant to the temporal scale of the fault. Under non-stationary fault arrival a model-derived dwell time reacts about four times faster than fixed hysteresis, 1.6 against 6.6 frames from onset to reconfiguration, at the cost of roughly 30\% more switching. Under stationary arrival the two are indistinguishable, so the auto-tuned dwell matches hand-tuning without the manual step.

\section{Does the Coupling Occur on Its Own?}
\label{sec:coupling}

Everything in Section~\ref{sec:results} is measured under a coupling we imposed. Whether real data already contains one is a separate question, and we tested it on RADIATE with four added sequences (\texttt{rain\_2\_0}, \texttt{rain\_3\_0}, \texttt{fog\_8\_0}, \texttt{fog\_8\_1}) on top of the four committed ones, eight in total. Table~\ref{tab:coupling} summarizes five checks.

\textbf{Measured coupling.} For every sequence we recorded real per-frame latency and detection count, quantities that do not depend on the fault signals, and re-ran the headline experiment with contention scheduled from the measured proxy in place of the fault label. Every correlation interval sits near zero or on the wrong side of it, and the de-circularized reduction collapses to about 0.0 pp on all eight sequences and both proxies. The imposed regime constructs the coupled scenario; by these proxies, RADIATE does not already contain one.

\textbf{An independent workload, corrected for multiple testing.} We added a third proxy, an ORB keypoint-matching workload that is CPU-bound and independent of the detector. Across 8 sequences $\times$ 3 proxies we computed paired two-sided $p$-values directly from the five seed-level differences of each test and applied Benjamini-Hochberg at $q$ = 0.05 over all 24. The tests are not independent, since proxies share frames within a sequence and sequences share a condition, but that is a positive-dependence structure under which the procedure still controls the false-discovery rate. One test came back nominally positive on the first run (\texttt{rain\_2\_0}, ORB proxy: +2.73 pp, $p$ $\approx$ 0.0024), narrowly missing its rank-1 threshold of 0.05/24 $\approx$ 0.0021. The ORB proxy is timed in wall-clock and so varies between runs, so we re-measured it, and on replication the effect vanished (+0.00 pp, $p$ = 1). After correction and replication, zero of 24 tests support naturally occurring coupling.

\textbf{Cross-sequence consistency.} For rain and fog, three sequences each, the per-sequence effect is positive in all six traces and the cluster bootstrap stays positive, but the cross-sequence $t$-interval includes zero for both conditions. Within a condition at $n$ = 3, the effect is directionally consistent without reaching significance. Pooling all 10 sequences across 6 conditions is a weaker and distinct claim, and it does clear the bar: mean +3.72 pp, $t$-interval [1.94, 5.50], cluster bootstrap [2.41, 5.43], sign test 10/10 positive ($p$ = 0.0010).

\textbf{Transfer of the whole belief model.} Fit the entire fault-belief estimator and $\kappa$ on one sequence and test on a held-out one, and the transferred detector still recovers faults (balanced accuracy 0.65 to 0.90, against 0.73 to 0.95 native). The routing benefit stays positive in all eight tested pairs and is significant in five, including transfers across weather conditions (rain to fog, +7.8 pp; fog to fog, +17.9 pp).

\textbf{Real annotations against pseudo-GT.} We projected RADIATE's radar-derived object annotations into the camera image and computed real, class-agnostic detection F1 for all four configurations on 500 rain frames. Real-GT F1 is far lower and far flatter across configurations, 0.19 to 0.22, than pseudo-GT F1 at 0.73 to 1.00. The configuration ranking does not match, and the per-frame correlation between the two is weak and sign-mixed. Coarser radar-projected boxes explain part of the absolute gap, but the ranking mismatch limits pseudo-GT to an internal consistency check. Every accuracy and utility number in this paper inherits that limitation. The deadline-miss results in Table~\ref{tab:rqh} come from timing alone and never touch an accuracy label, which leaves them unaffected.

\begin{table}[t]
\caption{Does the coupling occur naturally? Five checks on 8 RADIATE sequences.}
\label{tab:coupling}
\centering
\footnotesize
\setlength{\tabcolsep}{3pt}
\begin{tabular}{@{}>{\raggedright\arraybackslash}p{2.0cm}>{\raggedright\arraybackslash}p{5.8cm}@{}}
\toprule
\textbf{Check} & \textbf{Result} \\
\midrule
Measured coupling & Null: de-circularized $\Delta$ $\approx$ 0.0 pp against 1.1 to 9.4 pp imposed \\
\addlinespace[1pt]
Independent workload, BH-corrected & Null in 24/24 after correction; the one nominal positive failed replication \\
\addlinespace[1pt]
Cross-sequence generalization & Within condition ($n$ = 3): $t$-interval includes 0. Pooled over 10 sequences and 6 conditions: [1.94, 5.50], sign test $p$ = 0.001 \\
\addlinespace[1pt]
Full-model transfer & HMMs and $\kappa$ trained on another sequence: significant in 5/8 pairs, including across conditions \\
\addlinespace[1pt]
Real-GT accuracy & Configuration ranking does not match pseudo-GT; pseudo-GT is an internal check only \\
\bottomrule
\end{tabular}
\end{table}

After correction and replication we find no confirmed case of naturally occurring fault-contention coupling in un-manipulated RADIATE data. Section~\ref{sec:results} is therefore a proof of mechanism under imposed coupling. It says nothing about whether such coupling, or the accuracy differences it is scored against, arises in the field.

\section{Discussion}

The routing result holds across every dataset and camera geometry we tried. At the condition level the benefit tracks fault severity, by which we mean how often a fault makes the reference configuration infeasible; fault frequency and correlation strength predict it poorly. Since a threshold rule matches the noisy-OR, the supported claim is narrow: sensor evidence used to anticipate contention is what helps, and the fusion algebra is incidental. In ordinary data we could not find the coupling at all, in zero of 24 corrected tests.

The null bounds the method's downside. With no coupling present, the coupled router degrades exactly to the decoupled baseline: every uncoupled control and the fault-free trajectory sit at 0.00 pp, and $\kappa$ = 0 is exactly zero on all seven tracks. A perception stack can ship the router either way. The upside appears when coupling exists, and the standing cost is tens of microseconds per frame. Two things survive the null: the mechanism, which pays off wherever coupling is present, and a released evaluation pipeline (the imposed-schedule substrate, the de-circularization protocol, the independent-workload probe, and the multiple-testing procedure) that a deployment can run on its own traces to find out which regime it is in.

\textbf{Limitations.} The evaluation is offline and trace-driven, with a two-state contention model that ignores thermal effects, on one machine and not a shared embedded perception/planning/control stack. Pseudo-GT is only weakly validated against real annotations, and it also drives configuration ranking, so the accuracy and utility conclusions are provisional; the deadline-miss result fixes the timing mechanism without showing that the selected configurations are optimal under real accuracy. The five seeds vary only injected measurement noise on identical real frames, so seed intervals measure sensitivity to that noise on one trajectory. RADIATE frames are rectified left-camera only, and WoodScape's temporal axis is dataset order without continuous video. The natural next steps are a Jetson-class study with real embedded contention and more sequences to power the corrected coupling test. A closed-loop evaluation would then show whether any of this changes what the robot actually does.

\section{Conclusion}

We presented a perception router that couples a sensor-fault belief to a compute-contention belief through a noisy-OR term and uses the result to choose a detector configuration that fits the deadline. Where the coupling is present, the coupled policy lowers the deadline-miss rate by 1.1 to 9.4 pp, significant in five of six conditions and pooled over 10 sequences at $p$ = 0.001, at a routing cost of tens of microseconds per frame. A threshold rule on the same coupled features does just as well, which puts the weight of the result on the coupled features themselves. Five further checks probe the coupling regime instead of assuming it, and after correction and replication none of 24 tests find naturally occurring coupling in eight real sequences, though the belief model still transfers across sequences in five of eight held-out pairs. That leaves a proof of mechanism under imposed coupling. Whether such coupling occurs in the field is still open, and the released pipeline lets any deployment check its own traces.

\section*{Use of AI Assistance}

We used AI-based tools (large language model writing and coding assistants) while preparing this work. They helped improve the wording of the manuscript, worked through some technical and analysis decisions with us, and helped with minor debugging and code fixes in the released pipeline. The research questions, experimental design, implementation, data analysis, and results are our own. We reviewed and verified every AI-assisted passage and code change, and we are responsible for everything in this paper.

\end{document}